# $L_2$ Regularization for Learning Kernels

**Corinna Cortes**
Google Research
New York
corinna@google.com

**Mehryar Mohri**
Courant Institute and
Google Research
mohri@cims.nyu.edu

**Afshin Rostamizadeh**
Courant Institute
New York University
rostami@cs.nyu.edu

## Abstract

The choice of the kernel is critical to the success of many learning algorithms but it is typically left to the user. Instead, the training data can be used to learn the kernel by selecting it out of a given family, such as that of non-negative linear combinations of $p$ base kernels, constrained by a trace or $L_1$ regularization. This paper studies the problem of learning kernels with the same family of kernels but with an $L_2$ regularization instead, and for regression problems. We analyze the problem of learning kernels with ridge regression. We derive the form of the solution of the optimization problem and give an efficient iterative algorithm for computing that solution. We present a novel theoretical analysis of the problem based on stability and give learning bounds for orthogonal kernels that contain only an additive term $O(\sqrt{p/m})$ when compared to the standard kernel ridge regression stability bound. We also report the results of experiments indicating that $L_1$ regularization can lead to modest improvements for a small number of kernels, but to performance degradations in larger-scale cases. In contrast, $L_2$ regularization never degrades performance and in fact achieves significant improvements with a large number of kernels.

## 1 Introduction

Kernel methods have been successfully used in a variety of learning tasks (Schölkopf & Smola, 2002; Shawe-Taylor & Cristianini, 2004) with the best known example of support vector machines (SVMs) (Boser et al., 1992; Cortes & Vapnik, 1995; Vapnik, 1998). Positive definite symmetric (PDS) kernels specify an inner product in an implicit Hilbert space where large-margin methods are used for learning and estimation.

The choice of the kernel is critical to the success of the algorithm but in standard frameworks it is left to the user. A weaker commitment can be required from the user when instead the kernel is *learned* from data. One can then specify a family of kernels and let a learning algorithm use the data to select both the kernel out of this family and determine the prediction hypothesis.

The problem of learning kernels has been investigated in a number of recent publications including (Lanckriet et al., 2004; Micchelli & Pontil, 2005; Argyriou et al., 2005; Argyriou et al., 2006; Srebro & Ben-David, 2006; Ong et al., 2005; Lewis et al., 2006; Zien & Ong, 2007; Jebara, 2004; Bach, 2008). Some of this previous work examines families of Gaussian kernels (Micchelli & Pontil, 2005) or hyperkernels (Ong et al., 2005). But, the most common family of kernels considered is that of non-negative combinations of some fixed kernels constrained by a trace condition, which can be viewed as an $L_1$ regularization.

This paper studies the problem of learning kernels with the same family of kernels but with an $L_2$ regularization instead. Our analysis focuses on the regression setting also examined by Micchelli and Pontil (2005) and Argyriou et al. (2005). More specifically, we will consider the problem of learning kernels in kernel ridge regression, KRR, (Saunders et al., 1998). Our study is motivated by experiments carried out with a number of datasets, including those used by previous authors (Lanckriet et al., 2004; Cortes et al., 2008), in some of which using an $L_2$ regularization turned out to be significantly beneficial and otherwise never worse than using $L_1$ regularization. We report some of these results in the experimental section.

We also give a novel theoretical analysis of the problem of learning kernels in this context. A theoretical study of the problem of learning kernels in classification was previously presented by Srebro and Ben-David (2006) for SVMs and other similar classification algorithms. These authors proved that previous bounds given by Lanckriet et al. (2004) and Bousquet and Herrmann (2002) for the problem of learning kernels were vacuous. They further gave novel generalization bounds which, for linear combinations of kernels with $L_1$ regularization, have the form $R(h) \leq \widehat{R}(h) + \tilde{O}(\sqrt{(p + 1/\rho^2)/m})$, where $R(h)$ is the true error of a classifier $h$, $\widehat{R}(h)$ its empirical error, $p$ the number of kernels combined, $m$ the sample size, and $\rho$ the margin of the learned classifier (the notation $\tilde{\mathcal{O}}$ hides logarithmic factors in its arguments). Since the standard bound for SVMs has the form $R(h) \leq \widehat{R}(h) + \tilde{\mathcal{O}}(\sqrt{1/\rho^2)/m})$, this suggests that, up to logarithmic factors, the complexity term of the bound is only augmented with an additive term varying with $p$, in contrast with the multiplicative factor appearing in previous bounds, e.g., that of Micchelli and Pontil (2005) for the family of Gaussian kernels.

We give novel learning bounds with similar favorable guarantees for KRR with $L_2$ regularization. The complexity term of our bound as a function of $m$ and $p$ is of the form $O(1/\sqrt{m} + \sqrt{p/m})$ and is therefore only augmented by an additive term $O(\sqrt{p/m})$ with respect to the standard stability bound for KRR, with no additional logarithmic factor. Our bound is proven for the case where the base kernels are *orthogonal*. This assumption holds for the experiments in which the $L_2$ regularization yields significantly better results than $L_1$ but a similar, perhaps slightly weaker bound, is likely to hold in the general case. Our bound is based on a careful stability analysis of the algorithm for learning kernels with ridge regression and thus directly relates to the problem of learning kernels. A by-product of our analysis is a somewhat tighter stability bound and thus generalization bound for KRR.

The next two sections describe the optimization problem for learning kernels with ridge regression and give the form of its solution. We then present our stability analysis and generalization bound, leaving to the appendix much of the technical details. The last section briefly describes an iterative algorithm for determining the solution of regression learning problem that proved efficient in our experiments and reports the results of our experiments with a number of different datasets.

## 2 Optimization Problem

Let $S = ((x_1, y_1), \ldots, (x_m, y_m))$ denote the training sample and $\mathbf{y} = [y_1, \ldots, y_m]^\top$ the vector of training set labels, where $(x_i, y_i) \in X \times \mathbb{R}$ for $i \in [1, m]$, and let $\Phi(x)$ denote the feature vector associated to $x \in X$. Then, in the primal, the KRR optimization problem has the following form

$$\min_w \|w\|^2 + \frac{C}{m} \sum_{i=1}^m (w^\top \Phi(x_i) - y_i)^2, \tag{1}$$

where $C \geq 0$ is a trade-off parameter. For a fixed positive definite kernel (PDS) function $K \colon X \times X \to R$, the dual of the KRR optimization problem (Saunders et al., 1998) is given by:

$$\max_{\boldsymbol{\alpha}} -\lambda \boldsymbol{\alpha}^\top \boldsymbol{\alpha} - \boldsymbol{\alpha}^\top \mathbf{K} \boldsymbol{\alpha} + 2\boldsymbol{\alpha}^\top \mathbf{y}, \tag{2}$$

where $\mathbf{K} = (K(x_i, x_j))_{1 \leq i,j \leq m}$ is the Gram matrix associated to $K$ and where $\lambda = m/C$. In the following, we will denote by $\lambda_0$ the inverse of $C$, thus, $\lambda = \lambda_0 m$.

The idea of learning kernels is based on the principle of structural risk minimization (SRM) (Vapnik, 1998). It consists of selecting out of increasingly powerful kernels and thus hypothesis sets $H$, the one minimizing the minimum of a bound on the test error defined over $H$. Here, we limit the search to kernels $K$ that are non-negative combinations of $p$ fixed PDS kernels $K_k$, $k \in [1, p]$, and that are thereby guaranteed to be PDS, with an $L_2$ regularization: $\mathcal{K} = \{\sum_{k=1}^p \mu_k K_k \colon \boldsymbol{\mu} \in \mathcal{M}\}$, where $\mathcal{M} = \{\boldsymbol{\mu} \colon \boldsymbol{\mu} \geq 0 \wedge \|\boldsymbol{\mu} - \boldsymbol{\mu}_0\|^2 \leq \Lambda^2\}$, with $\boldsymbol{\mu} = [\mu_1, \ldots, \mu_p]^\top$, $\boldsymbol{\mu}_0 > 0$ a fixed combination vector, and $\Lambda \geq 0$ a regularization parameter. In view of the multiplier $\Lambda$, we can assume, without loss of generality, that the minimum component of $\boldsymbol{\mu}_0$ is one.

Based on the dual form of the optimization problem for KRR, the kernel learning optimization problem can be formulated as follows:

$$\min_{\boldsymbol{\mu} \in \mathcal{M}} \max_{\boldsymbol{\alpha}} -\lambda \boldsymbol{\alpha}^\top \boldsymbol{\alpha} - \underbrace{\sum_{k=1}^p \mu_k \boldsymbol{\alpha}^\top \mathbf{K}_k \boldsymbol{\alpha}}_{\boldsymbol{\mu}^\top \mathbf{v}} + 2\boldsymbol{\alpha}^\top \mathbf{y}, \tag{3}$$

where $\mathbf{K}_k$ is the Gram matrix associated to the base kernel $K_k$. It is convenient to introduce the vector $\mathbf{v} = [v_1, \ldots, v_p]^\top$ where $v_k = \boldsymbol{\alpha}^\top \mathbf{K}_k \boldsymbol{\alpha}$. Note that this defines a convex optimization problem in $\boldsymbol{\mu}$, since the objective function is linear in $\boldsymbol{\mu}$ and the pointwise maximum over $\boldsymbol{\alpha}$ preserves convexity, and since $\mathcal{M}$ is a convex set. We refer in short by LKRR to this learning kernel KRR procedure and denote by $h$ the hypothesis it returns defined by $h(x) = \sum_{i=1}^m \alpha_i K(x_i, x)$ for all $x \in X$, when trained on the sample $S$, where $K$ denotes the PDS kernel $K = \sum_{k=1}^p \mu_k K_k$.

## 3 Form of the Solution

**Theorem 1.** *The solution $\boldsymbol{\mu}$ of the optimization problem (3) is given by $\boldsymbol{\mu} = \boldsymbol{\mu}_0 + \Lambda \frac{\mathbf{v}}{\|\mathbf{v}\|}$ with $\boldsymbol{\alpha}$ the unique vector verifying $\boldsymbol{\alpha} = (\mathbf{K} + \lambda \mathbf{I})^{-1} \mathbf{y}$.*

*Proof.* By von Neumann's (1937) generalized minimax theorem, (3) is equivalent to its max-min analogue:

$$\max_{\boldsymbol{\alpha}} -\lambda \boldsymbol{\alpha}^\top \boldsymbol{\alpha} + 2\boldsymbol{\alpha}^\top \mathbf{y} + \min_{\boldsymbol{\mu} \in \mathcal{M}} -\boldsymbol{\mu}^\top \mathbf{v}, \tag{4}$$

where $\mathbf{v} = (\boldsymbol{\alpha}^\top K_1 \boldsymbol{\alpha}, \ldots, \boldsymbol{\alpha}^\top K_p \boldsymbol{\alpha})^\top$. The Lagrangian of the minimization problem is $L = -\boldsymbol{\mu}^\top(\mathbf{v} + \boldsymbol{\beta}) + \gamma(\|\boldsymbol{\mu} - \boldsymbol{\mu}_0\|^2 - \Lambda^2)$ with $\boldsymbol{\beta} \geq 0$ and $\gamma \geq 0$ and the KKT conditions are

$$\nabla_{\boldsymbol{\mu}} L = -(\mathbf{v} + \boldsymbol{\beta}) + 2\gamma(\boldsymbol{\mu} - \boldsymbol{\mu}_0) = 0$$

$$\nabla_{\boldsymbol{\beta}} L = \boldsymbol{\mu}^\top \boldsymbol{\beta} = 0 \Rightarrow (\frac{\mathbf{v} + \boldsymbol{\beta}}{2\gamma} + \boldsymbol{\mu}_0)^\top \boldsymbol{\beta} = 0$$

$$\gamma(\|\boldsymbol{\mu} - \boldsymbol{\mu}_0\|^2 - \Lambda^2) = 0.$$

Note that if $\gamma = 0$ then the $L_2$ constraint is not met as an equality, which cannot hold at the optimum. By inspecting (3), it is clear that the $\mu_k$s would be chosen as large as possible. Thus, the first equality implies $\boldsymbol{\mu} - \boldsymbol{\mu}_0 = \frac{\mathbf{v}+\boldsymbol{\beta}}{2\gamma}$, in view of which the second gives $-\|\boldsymbol{\beta}\|^2 = (\frac{\mathbf{v}}{2\gamma} + \boldsymbol{\mu}_0)^\top \boldsymbol{\beta}$. Since $\mathbf{v} \geq 0, \boldsymbol{\mu}_0 \geq 0, \gamma \geq 0$ and $\boldsymbol{\beta} \geq 0$, $(\frac{\mathbf{v}}{2\gamma} + \boldsymbol{\mu}_0)^\top \boldsymbol{\beta}$ is non-negative, which implies $-\|\boldsymbol{\beta}\|^2 \geq 0$ and $\boldsymbol{\beta} = 0$. The third equality gives $\boldsymbol{\mu} - \boldsymbol{\mu}_0 = \Lambda \frac{\mathbf{v}}{\|\mathbf{v}\|}$. Problem 4 can thus be rewritten as

$$\max_{\boldsymbol{\alpha}} \underbrace{-\lambda \boldsymbol{\alpha}^\top \boldsymbol{\alpha} + 2\boldsymbol{\alpha}^\top \mathbf{y} - \boldsymbol{\mu}_0^\top \mathbf{v}}_{\text{standard KRR with } \boldsymbol{\mu}_0\text{-kernel } \mathbf{K}_0.} - \Lambda \|\mathbf{v}\|. \tag{5}$$

For $\mathbf{v} \neq 0$, $\nabla_{\boldsymbol{\alpha}} \|\mathbf{v}\| = 2 \sum_{k=1}^p \frac{v_k}{\|\mathbf{v}\|} \mathbf{K}_k \boldsymbol{\alpha}$. Thus, differentiating and setting to zero the objective function of this optimization problem gives $\boldsymbol{\alpha} = (\mathbf{K} + \lambda \mathbf{I})^{-1} \mathbf{y}$, with $\mathbf{K} = \sum_{k=1}^p \Big( \boldsymbol{\mu}_{0k} + \Lambda \frac{v_k}{\|\mathbf{v}\|}_{\mu_k} \Big) \mathbf{K}_k = \sum_{k=1}^p \mu_k \mathbf{K}_k$. □

## 4 Stability analysis

We will derive generalization bounds for LKRR using the notion of algorithmic stability (Bousquet & Elisseeff, 2002). A learning algorithm is said to be (uniformly) $\beta$-*stable* if the hypotheses $h'$ and $h$ it returns for any two training samples, $S$ and $S'$, that differing by a single point satisfy $\big|[h'(x) - y]^2 - [h(x) - y]^2\big| \leq \beta$ for any point $x \in X$ labeled with $y \in \mathbb{R}$. The stability coefficient $\beta$ is a function of the sample size $m$. Stability in conjunction with McDiarmid's inequality can lead to tight generalization bounds specific to the algorithm analyzed (Bousquet & Elisseeff, 2002).

We analyze the stability of LKRR. Thus, we consider two samples of size $m$, $S$ and $S'$, differing only by $(x_m, y_m)$ ($(x'_m, y'_m)$ in $S'$) and bound the $|h'(x) - h(x)|$. The analysis is quite complex in this context and the standard convexity-based proofs of Bousquet and Elisseeff (2002) do not readily apply. This is because here, a change in a sample point also changes the PDS kernel $K$, which in the standard case is fixed.

Our proofs are novel and make use of the expression of $\boldsymbol{\alpha}$ and $\boldsymbol{\mu}$ supplied by Theorem 1, which can lead to tighter bounds. In particular, the same analysis gives us a novel and somewhat tighter bound on the stability of standard KRR than the one obtained via convexity arguments (Bousquet & Elisseeff, 2002).

Fix $x \in X$. We shall denote by $\Delta h(x)$ the difference $h'(x) - h(x)$ and more generally use the symbol $\Delta$ to abbreviate the difference between an expression depending on $S'$ and one depending on $S$. We derive a bound on $\Delta h(x) = h'(x) - h(x)$ for LKRR. We denote by $\mathbf{y}'$ the vector of labels, by $K'$ the kernel learned by LKRR, and by $\mu'_k$ and $\boldsymbol{\mu}'$ the basis kernel coefficients and vector associated to the sample $S'$.

We will assume that the hypothesis set considered is bounded, that is $|h(x) - y(x)| \leq M$ for all $x \in X$, for some $M \geq 0$. This bound and the Lipschitz property of the loss function implies a bound on $\Delta(h(x) - y)^2 \leq 2M \Delta h(x)$. We will also assume that the base kernels are bounded: there exists $\kappa_0 \geq 0$ such that $(\sum_{k=1}^p K_k(x,x)^2)^{1/2} \leq \kappa_0$ for all $x \in X$. This implies that for all $x \in X$, $K(x,x) = \sum_{k=1}^p \mu_k K_k(x,x) \leq \kappa_0 \|\boldsymbol{\mu}\| \leq \kappa_0(\|\boldsymbol{\mu}_0\| + \Lambda)$. Thus, we can assume that there exists $\kappa \geq 0$ such that $K(x,x) \leq \kappa$ for all $x \in X$.

Now, $\Delta h(x)$ can be written as $\Delta h(x) = \Delta_S h(x) + \Delta_K h(x)$ to distinguish changes due to different samples ($x'_i$s vs $x_i$s) for a fixed kernel and those due to a different kernels $K$ for a fixed sample:

$$\begin{aligned}
\Delta_S h(x) = &\sum_{i=1}^m \Big[ (\sum_{k=1}^p \mu'_k \mathbf{K}_k(S') + \lambda \mathbf{I})^{-1} \mathbf{y}' \Big]_i \sum_{k=1}^p \mu'_k K_k(x'_i, x) \\
&- \sum_{i=1}^m \Big[ (\sum_{k=1}^p \mu'_k \mathbf{K}_k(S) + \lambda \mathbf{I})^{-1} \mathbf{y} \Big]_i \sum_{k=1}^p \mu'_k K_k(x_i, x), \\
\Delta_K h(x) = &\sum_{i=1}^m \Big[ (\sum_{k=1}^p \mu'_k \mathbf{K}_k(S) + \lambda \mathbf{I})^{-1} \mathbf{y} \Big]_i \sum_{k=1}^p \mu'_k K_k(x_i, x) \\
&- \sum_{i=1}^m \Big[ (\sum_{k=1}^p \mu_k \mathbf{K}_k(S) + \lambda \mathbf{I})^{-1} \mathbf{y} \Big]_i \sum_{k=1}^p \mu_k K_k(x_i, x),
\end{aligned}$$

where $\mathbf{K}_k(S)$ (resp. $\mathbf{K}_k(S')$) is the kernel matrix generated from $S$ (resp. $S'$). We bound these two terms separately. The main reason for this is that the term $\Delta_S h(x)$ leads to sparse expressions since the points $x_i$s in $S$ and $S'$ differ only by $x_m$ and $x'_m$. To bound $\Delta_K h(x)$ other techniques are needed.

In what follows, we denote by $\Phi$ a feature mapping associated to kernel $K$ and by $\boldsymbol{\Phi}$ the matrix whose columns are $\Phi(x_i)$, $i = 1, \ldots, m$. Similarly, we denote by $\boldsymbol{\Phi}'$ the matrix whose columns are $\Phi(x'_i)$, $i = 1, \ldots, m$, and for $k = 1, \ldots, p$, we denote by $\Phi_k$ a feature mapping associated with the base kernel $K_k$ and by $\boldsymbol{\Phi}_k$ the matrix whose columns are $\Phi_k(x_i)$, $i = 1, \ldots, m$.

### 4.1 Bound on $\Delta_S h(x)$

For the analysis of $\Delta_S h(x)$, the kernel coefficients $\mu'_k$ are fixed. Here, we denote by $\mathbf{K}$ the kernel matrix of $\sum_{k=1}^p \mu'_k \mathbf{K}_k$ over the sample $S$, and by $\mathbf{K}'$ the one over $S'$. Now, $h(x)$ can be expressed in terms of $\boldsymbol{\Phi}$ as follows:

$$\begin{aligned}
h(x) &= [\boldsymbol{\Phi}\boldsymbol{\alpha}]^\top \Phi(x) = \mathbf{y}^\top (\mathbf{K} + \lambda \mathbf{I})^{-1} \boldsymbol{\Phi}^\top \Phi(x) && (6) \\
&= \mathbf{y}^\top (\boldsymbol{\Phi}^\top \boldsymbol{\Phi} + \lambda \mathbf{I})^{-1} \boldsymbol{\Phi}^\top \Phi(x). && (7)
\end{aligned}$$

**Theorem 2.** *Let $\lambda_{\min}$ denote the smallest eigenvalue of $\boldsymbol{\Phi}'\boldsymbol{\Phi}'^\top$. Then, the following bound holds for all $x \in X$:*

$$|\Delta_S h(x)| \leq \frac{2\kappa M}{\lambda_{\min} + \lambda_0 m}. \tag{8}$$

*Proof.* Using the general identity $(\mathbf{\Phi}^\top\mathbf{\Phi} + \lambda\mathbf{I})^{-1}\mathbf{\Phi}^\top = \mathbf{\Phi}^\top(\mathbf{\Phi}\mathbf{\Phi}^\top + \lambda\mathbf{I})^{-1}$, we can write equation (7) as

$$h(x) = (\mathbf{\Phi}\mathbf{y})^\top(\mathbf{\Phi}\mathbf{\Phi}^\top + \lambda\mathbf{I})^{-1}\Phi(x). \tag{9}$$

Let $\mathbf{U} = (\mathbf{\Phi}\mathbf{\Phi}^\top + \lambda\mathbf{I})$ and denote by $\mathbf{w}^\top$ the row vector $(\mathbf{\Phi}\mathbf{y})^\top\mathbf{U}^{-1}$. Now, we can write $\Delta_S h(x) = (\Delta_S\mathbf{w})^\top\Phi'(x)$. Using the identity $\Delta_S(\mathbf{U}^{-1}) = -\mathbf{U}^{-1}(\Delta_S\mathbf{U})\mathbf{U}'^{-1}$, valid for all invertible matrices $\mathbf{U}$ and $\mathbf{U}'$, $\Delta_S\mathbf{w}^\top$ can be expressed as follows:

$$\begin{aligned}\Delta_S\mathbf{w}^\top &= (\Delta_S\mathbf{\Phi}\mathbf{y})^\top\mathbf{U}'^{-1} + (\mathbf{\Phi}\mathbf{y})^\top\Delta_S(\mathbf{U}^{-1})\\ &= (\Delta_S\mathbf{\Phi}\mathbf{y})^\top\mathbf{U}'^{-1} - (\mathbf{\Phi}\mathbf{y})^\top\mathbf{U}^{-1}(\Delta_S\mathbf{U})\mathbf{U}'^{-1}.\end{aligned}$$

We observe that

$$\begin{aligned}(\Delta_S\mathbf{\Phi}\mathbf{y}) &= \Delta_S(\sum_{i=1}^m y_i\Phi(x_i)) = \sum_{i=1}^m(\Delta_S y_i\Phi(x_i))\\ &= \Delta_S(y_m\Phi(x_m)) \qquad \text{and}\\ (\Delta_S\mathbf{U}) &= \Delta_S(\sum_{i=1}^m\Phi(x_i)\Phi(x_i)^\top) = \Delta_S(\Phi(x_m)\Phi(x_m)^\top).\end{aligned}$$

Thus, we can write $\Delta_S\mathbf{w}^\top$

$$\begin{aligned}&= \Big[\Delta_S(y_m\Phi(x_m))^\top - (\mathbf{\Phi}\mathbf{y})^\top\mathbf{U}^{-1}\Delta_S(\Phi(x_m)\Phi(x_m)^\top)\Big]\mathbf{U}'^{-1}\\ &= \Big[y'_m\Phi(x'_m)^\top - y_m\Phi(x_m)^\top + (\mathbf{\Phi}\mathbf{y})^\top\mathbf{U}^{-1}\Phi(x'_m)\Phi(x'_m)^\top\\ &\quad - (\mathbf{\Phi}\mathbf{y})^\top\mathbf{U}^{-1}\Phi(x_m)\Phi(x_m)^\top\Big]\mathbf{U}'^{-1}\\ &= \Big[(y'_m - h(x'_m))\Phi(x'_m) - (y_m - h(x_m))\Phi(x_m)\Big]^\top\mathbf{U}'^{-1}.\end{aligned}$$

Since for all $x \in X$, $K(x,x) \leq \kappa$ and $|h(x) - y(x)| \leq M$, we have $\|\Phi(x)\| \leq \kappa^{1/2}$ and $\|(y'_m - h(x'_m))\Phi(x'_m) - (y_m - h(x_m))\Phi(x_m)\| \leq 2\kappa^{1/2}M$, thus

$$\|\Delta_S\mathbf{w}^\top\| \leq 2\kappa^{1/2}M\|\mathbf{U}'^{-1}\|. \tag{10}$$

The smallest eigenvalue of $(\mathbf{\Phi}'\mathbf{\Phi}'^\top + \lambda\mathbf{I})$ is $\lambda_{\min} + \lambda$. Thus, $\|\Delta_S\mathbf{w}^\top\| \leq \frac{2\kappa^{1/2}M}{\lambda_{\min}+\lambda_0 m}$. Since $\|\Phi'(x)\| = K'(x,x) \leq \kappa^{1/2}$, $|\Delta_S h(x)| \leq \frac{2\kappa M}{\lambda_{\min}+\lambda_0 m}$. □

Recall, $\Delta_S h(x)$ represents the variation due to sample changes for a fixed kernel, thus, the bound given by the theorem is precisely a bound on the stability coefficient of standard KRR. This bound is tighter than the one obtained using the techniques of Bousquet and Elisseeff (2002): $|\Delta_S h(x)| \leq \frac{2\kappa M}{\lambda_0 m}$. Also, since $\mathbf{\Phi}'\mathbf{\Phi}'^\top$ and $\mathbf{K}' = \mathbf{\Phi}'^\top\mathbf{\Phi}'$ have the same non-zero eigenvalues, when $\lambda_{\min} \neq 0$, $\lambda_{\min}$ is the smallest non-zero eigenvalue of $\mathbf{K}'$, $\lambda^*_{\min}(\mathbf{K}')$.

### 4.2 Bound on $\Delta_K h(x)$

Since $h(x) = \sum_{i=1}^m \alpha_i K(x_i, x)$, the variation in $K$ can be decomposed into the following sum:

$$\Delta_K h(x) = \underbrace{\sum_{i=1}^m(\Delta_K\alpha_i)K'(x'_i,x)}_{R} + \underbrace{\sum_{i=1}^m\alpha_i\Delta_K K(x'_i,x)}_{T}.$$

By the Cauchy-Schwarz inequality, for any $x'_i, x \in X$, $|K(x'_i,x)| \leq \sqrt{K(x'_i,x'_i)K(x,x)} \leq \kappa$, thus the norm of the vector $k_{x'} = [K(x'_1,x),\ldots,K(x'_m,x)]$ is bounded by $\kappa\sqrt{m}$ and the first term $R$ can be bounded straightforwardly in terms of $\Delta_K\boldsymbol{\alpha}$: $|R| \leq \kappa\sqrt{m}\|\Delta_K\alpha\|$.

The second term can be written as follows

$$T = \sum_{i=1}^m\alpha_i\sum_{k=1}^p(\Delta\mu_k)K_k(x'_i,x) = \sum_{k=1}^p(\Delta\mu_k)(\mathbf{\Phi}_k\boldsymbol{\alpha})^\top\Phi_k(x). \tag{11}$$

By Lemma 1 (see Appendix), $\Delta\mu_k$ can be expressed in terms of the $\Delta v_k$s and thus $T$ can be rewritten as

$$T = \underbrace{\Lambda\sum_{k=1}^p\left[\frac{\Delta v_k}{\|\mathbf{v}'\|} - \frac{v_k\sum_{i=1}^p(v_i+v'_i)\Delta v_i}{\|\mathbf{v}\|\|\mathbf{v}'\|(\|\mathbf{v}\|+\|\mathbf{v}'\|)}\right](\mathbf{\Phi}_k\boldsymbol{\alpha})^\top\,\Phi_k(x)}_{V}. \tag{12}$$

Note, in order to isolate the term $V$ each $\Phi_k$ must map to the same feature space. This holds for the empirical kernel map, or any orthogonal kernels as will be defined below. In this expression, each $\Delta v_k$ can be written as a sum $\Delta v_k = \Delta_K v_k + \Delta_S v_k$, where

$$\begin{aligned}\Delta_K v_k =&\,\mathbf{y}'^\top(\mathbf{K}'+\lambda\mathbf{I})^{-1}\mathbf{K}_k(S')(\mathbf{K}'+\lambda\mathbf{I})^{-1}\mathbf{y}' &(13)\\ &-\mathbf{y}'^\top(\mathbf{K}+\lambda\mathbf{I})^{-1}\mathbf{K}_k(S')(\mathbf{K}+\lambda\mathbf{I})^{-1}\mathbf{y}' &(14)\\ \Delta_S v_k =&\,\mathbf{y}'^\top(\mathbf{K}+\lambda\mathbf{I})^{-1}\mathbf{K}_k(S')(\mathbf{K}+\lambda\mathbf{I})^{-1}\mathbf{y}' &(15)\\ &-\mathbf{y}^\top(\mathbf{K}+\lambda\mathbf{I})^{-1}\mathbf{K}_k(S)(\mathbf{K}+\lambda\mathbf{I})^{-1}\mathbf{y}. &(16)\end{aligned}$$

Let $V = V_1 + V_2$ where $V_1$ (resp. $V_2$) is the expression corresponding to $\Delta_K$ (resp. $\Delta_S$). We will denote by $V_k$, $V_{1k}$ and $V_{2k}$ each of the terms depending on $k$ appearing in their sum. The proof of the propositions giving bounds on $\|V_1\|$ and $\|V_2\|$ are left to the appendix.

**Proposition 1.** *For any samples $S$ and $S'$ differing by one point, the following inequality holds:*

$$\|V_1\| \leq 4\Lambda\sqrt{\kappa pm}\,\|\Delta_K\boldsymbol{\alpha}\|. \tag{17}$$

Our bound on $V_2$ holds for *orthogonal* base kernels.

**Definition 1.** *Kernels $K_1,\ldots,K_k$ are said to be* orthogonal *if they admit feature mappings $\Phi_k\colon X \mapsto F$ mapping to the same Hilbert space $F$ such that for all $x \in X$, and $i \neq j$,*

$$\Phi_i(x)^\top\Phi_j(x) = 0. \tag{18}$$

This assumption is satisfied in particular by the n-gram based kernels used in our experiments and more generally by kernels $K_k$ whose feature mapping can be obtained by projecting the feature vector $\Phi(x)$ of some kernel $K$ on orthogonal spaces. The *concatenation* type kernels suggested by Bach (2008) are also a special case of orthogonal kernels.

**Proposition 2.** *Assume that the base kernels $K_k$, $k \in [1,p]$ are orthogonal. Then, for any samples $S$ and $S'$ differing*

*by one point, the following inequality holds:*

$$\|V_2\| \leq \frac{4\Lambda M}{\lambda_0 m}. \tag{19}$$

Combining the bounds on $V_1$ and $V_2$ gives

$$\|V\| \leq 4\Lambda\sqrt{\kappa p m}\,\|\Delta_K \boldsymbol{\alpha}\| + \frac{4\Lambda M}{\lambda_0 m}.$$

$\Delta_K \boldsymbol{\alpha} = -(\mathbf{K}' + \lambda \mathbf{I})^{-1}(\Delta \mathbf{K})\boldsymbol{\alpha}$ can be expressed in terms of the $V_k$s as follows:

$$\Delta_K \boldsymbol{\alpha} = -(\mathbf{K}' + \lambda \mathbf{I})^{-1} \sum_{k=1}^{p} (V_k \boldsymbol{\Phi}_k)^\top.$$

Decomposing $V_k$ as in $V_k = V_{1k} + V_{2k}$, using the expression of $V_{1k}$ from (24), and collecting all $\Delta_K \boldsymbol{\alpha}$ terms to the left hand side, leads to the following expression relating $\Delta_K \boldsymbol{\alpha}$ to the $V_{2k}$s:

$$\Delta_K \boldsymbol{\alpha} = -\mathbf{Y}^{-1}\Big(\sum_{k=1}^{p} (V_{2k} \boldsymbol{\Phi}_k)^\top\Big), \tag{20}$$

with $\mathbf{Y} = \mathbf{K}' + \lambda \mathbf{I} + \Lambda \sum_{k=1}^{p} \frac{\mathbf{K}_k}{\|\mathbf{v}'\|} \boldsymbol{\alpha}\boldsymbol{\alpha}^\top \mathbf{Q}_k$, and $\mathbf{Q}_k = \big[\mathbf{K}_k - \frac{v_k}{\|\mathbf{v}\|} \frac{\sum_{i=1}^{p}(v_i + v_i')\mathbf{K}_i}{\|\mathbf{v}\| + \|\mathbf{v}'\|}\big]$. $\boldsymbol{\alpha}\boldsymbol{\alpha}^\top \mathbf{Q}_k$ has rank one since $\boldsymbol{\alpha}\boldsymbol{\alpha}^\top$ is a projection on the line spanned by $\boldsymbol{\alpha}$ and its trace $\mathrm{Tr}[\boldsymbol{\alpha}\boldsymbol{\alpha}^\top \mathbf{Q}_k] = \boldsymbol{\alpha}^\top \mathbf{Q}_k \boldsymbol{\alpha}$ is non-negative:

$$\begin{aligned}\boldsymbol{\alpha}^\top \mathbf{Q}_k \boldsymbol{\alpha} &= v_k - \frac{v_k}{\|\mathbf{v}\|} \frac{\sum_{i=1}^{p}(v_i^2 + v_i' v_i)}{\|\mathbf{v}\| + \|\mathbf{v}'\|} \\ &\geq v_k - \frac{v_k}{\|\mathbf{v}\|} \frac{\|\mathbf{v}\|^2 + \|\mathbf{v}'\|\|\mathbf{v}\|}{\|\mathbf{v}\| + \|\mathbf{v}'\|} = v_k - v_k = 0,\end{aligned}$$

using the Cauchy-Schwarz inequality. Thus, the eigenvalues of $\boldsymbol{\alpha}\boldsymbol{\alpha}^\top \mathbf{Q}_k$ are non-negative and since it has rank one and $\mathbf{K}_k$ is positive-semidefinite, the eigenvalues of $\mathbf{K}_k \boldsymbol{\alpha}\boldsymbol{\alpha}^\top \mathbf{Q}_k$ are also non-negative. This implies that the smallest eigenvalue of $\mathbf{Y}$ is at least $\lambda$ and that $\|\mathbf{Y}^{-1}\| \leq 1/(\lambda_0 m)$. Since $\|\sum_{k=1}^{p} V_{2k} \Phi_k\| \leq \|V_2\|\sqrt{\kappa m}$, this leads to

$$\|V\| \leq \frac{4\Lambda M (4\Lambda \kappa p^{1/2}/\lambda_0 + 1)}{\lambda_0 m}, \tag{21}$$

and the following result.

**Proposition 3.** *The uniform stability of LKRR can be bounded as follows:*

$$|\Delta(h(x) - y)^2| \leq 2M|\Delta h(x)| \leq 2M \frac{C_0 + C_1\sqrt{p}}{\lambda_0 m}, \tag{22}$$

*with* $C_0 = 2\kappa M + 4\Lambda M \kappa^{1/2}(\kappa/\lambda_0 + 1)$ *and* $C_1 = 16\Lambda^2 M \kappa^{3/2}/\lambda_0$.

A direct application of the general stability bound (Bousquet & Elisseeff, 2002) or the application of McDiarmid's inequality yields the following generalization bound for LKRR.

**Theorem 3.** *Let $h$ denote the hypothesis returned by LKRR and assume that for for all $x \in X$, $|h(x) - y(x)| \leq M$. Then, for any $\delta > 0$, with probability at least $1 - \delta$,*

$$R(h) \leq \widehat{R}(h) + 2\beta + \big(4m\beta + M\big)\sqrt{\frac{\log\frac{1}{\delta}}{2m}},$$

*where $\beta = O(1/m) + O(\sqrt{p}/m)$ is the stability bound given by Proposition 3.*

Thus, in view of this theorem our generalization bound has the form $R(h) \leq \widehat{R}(h) + O(1/\sqrt{m} + \sqrt{p/m})$.

## 5 Experimental Results

In this section we examine the performance of $L_2$-regularized kernel-learning on a number of datasets.

Problem (5) is a convex optimization problem and can thus be solved using standard gradient descent-type algorithms. However, the form of the solution provided by Theorem 1, $\boldsymbol{\alpha} = (\mathbf{K} + \lambda \mathbf{I})^{-1}$, motivates an iterative algorithm that proved to be significantly faster in our experiments. The following gives the pseudocode of the algorithm, where $\eta \in (0, 1)$ is an interpolation parameter and $\epsilon > 0$ a convergence error. In our experiments, the number of iterations

**Algorithm 1** Interpolated Iterative Algorithm

**Input:** $\mathbf{K}_k$, $k \in [1, p]$
$\boldsymbol{\alpha}' \leftarrow (\mathbf{K}_0 + \lambda \mathbf{I})^{-1}\mathbf{y}$
**repeat**
  $\boldsymbol{\alpha} \leftarrow \boldsymbol{\alpha}'$
  $\mathbf{v} \leftarrow (\boldsymbol{\alpha}^\top K_1 \boldsymbol{\alpha}, \ldots, \boldsymbol{\alpha}^\top K_p \boldsymbol{\alpha})^\top$
  $\boldsymbol{\mu} \leftarrow \boldsymbol{\mu}_0 + \Lambda \frac{\mathbf{v}}{\|\mathbf{v}\|}$
  $\boldsymbol{\alpha}' \leftarrow \eta \boldsymbol{\alpha} + (1 - \eta)(\mathbf{K}(\boldsymbol{\alpha}) + \lambda \mathbf{I})^{-1}\mathbf{y}$
**until** $\|\boldsymbol{\alpha}' - \boldsymbol{\alpha}\| < \epsilon$

needed on average for convergence was about 10 to 15 with $\eta = 1/2$. When using a small number of kernels with few data points, each iteration took a fraction of a second, while when using thousands of kernels and data-points each iteration took about a second. In view of the space limitations, we do not present a bound on the number of iterations. But, it should be clear that bounding techniques similar to what we used for the stability analysis can be used to estimate the Lipschitz constant of the function $f\colon \boldsymbol{\alpha} \mapsto (\mathbf{K} + \lambda \mathbf{I})^{-1}\mathbf{y}$, which yields directly a bound on the number of iterations.

We did two series of experiments. First, we validated our experimental set-up and our implementation for Algorithm 1 and previous algorithms for L1 regularization by comparing our results against those previously presented by Lanckriet et al. (2004), which use a small number of base kernels and relatively small data sets. We then focused on a larger task consisting of learning sequence kernels using thousands of base kernels as described by Cortes et al. (2008).

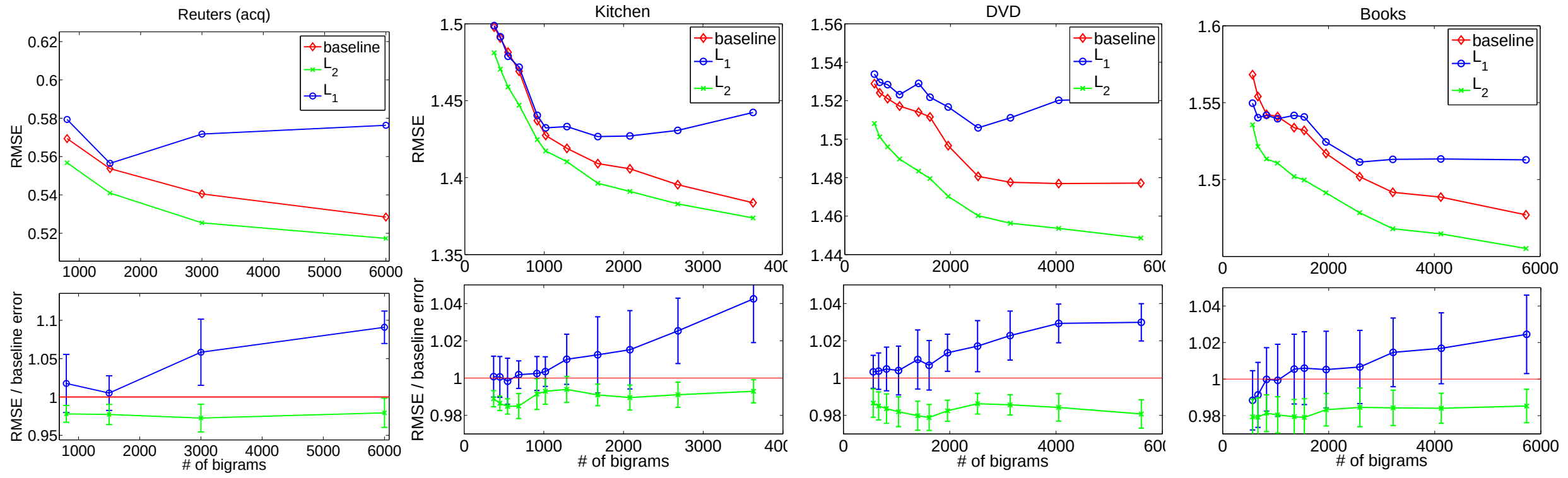

Figure 1: RMSE error reported for the Reuters and various sentiment analysis datasets (kitchen, DVDs and electronics). The upper plots show the absolute error, while the bottom plots show the error after normalizing by the baseline error (error bars are $\pm 1$ standard deviation).

### 5.1 UCI datasets

To verify our implementation, we first evaluated Algorithm 1 on the *breast*, *ionosphere*, *sonar* and *heart* datasets from the UCI ML Repository which were previously used for experimentation by Lanckriet et al. (2004). In order to use KRR for the classification datasets, we train with $\pm 1$ labels and examined both root mean squared error (RMSE) with respect to these target values and the misclassification rate when using the sign of the learned function to classify the test set. We found that both measures of error give similar comparative results. We use exactly the same experimental setup as (Lanckriet et al., 2004), with three kernels: a Gaussian, a linear, and a second degree polynomial kernel.

For comparison, we consider the best performing single kernel of these three kernels, the performance of an evenly-weighted sum of the kernels, and the performance of an $L_1$-regularized algorithm (similar to that of Lanckriet et al. (2004), however using the KRR objective).

Our results on these datasets validate our implementations by reaffirming the results from Lanckriet et al. (2004). Using kernel-learning algorithms (whether $L_1$ or $L_2$ regularized) never does worse than selecting the best single kernel via costly cross-validation. However, our experiments also confirm the findings by Lanckriet et al. (2004) that kernel-learning algorithms for this setting never do significantly *better*. All differences are easily within one standard deviation, with absolute misclassification rate of: 0.03 (*breast*), 0.08 (*ionosphere*), 0.16 (*sonar*) and 0.17 (*heart*). As our next set of experiments will show, when the number of base kernels is substantially increased, this picture changes completely. The performance of the $L_2$ regularized kernel is significantly better than the baseline of evenly-weighted sum of kernels, that in turn performs significantly better than the $L_1$ regularized kernel.

### 5.2 Sequence-based datasets

In our next experiments, we also make use of one of the datasets from (Lanckriet et al., 2004), the ACQ task of the Reuters-21578 dataset, though we learn with different base kernels. Using the ModApte split we produce 3,299 test examples and 9,603 training examples from which we randomly subsample 2,000 points to train with over 20 trials.

For features we use the $N$ most frequently occurring bigrams, where $N$ is indicated in Figure 1. As suggested in Cortes et al. (2008), we use $N$ rank-1 base kernels, with each kernel corresponding to a particular n-gram. Thus, if $\mathbf{v}_i \in \mathbb{R}^m$ is the vector of the occurrences of the $i$th n-gram across the training data, then the $i$th base kernel matrix is defined as $K_i = \mathbf{v}_i \mathbf{v}_i^\top$. As is common for KRR, we also include a constant feature, and thus kernel, which acts as an offset. Note that these base kernels are orthogonal, since each $\Phi_i$ is the projection onto a single distinct component of $\Phi$. The parameters $\lambda$ and $\Lambda$ are chosen via 10-fold cross validation on the training data.

We compare the presented $L_2$-regularized algorithm to both a baseline of the evenly-weighted sum of all the base kernels, as well as to the $L_1$-regularized method of Cortes et al. (2008) (Figure 1). The results illustrate that for large-scale kernel-learning, kernel selection with $L_2$ regularization improves performance, and that $L_1$ regularization can in fact be harmful. Note, that all base kernels here represent orthogonal features, thus, a sparse solution that eliminates a subset of the base kernels may negatively impact performance. Since Lanckriet et al. (2004) do not perform learning for large number of base kernels, we cannot directly compare results for this task. However, the best error rate we obtain by classifying the test set by the sign of the $L_1$-regularized learner is comparable to that reported by Lanckriet et al. (2004).

For our last experiments we consider the task of sentiment analysis of reviews within several domains: books, dvds, and kitchen appliances (Blitzer et al., 2007). Each domain consists of 2,000 product reviews, each with a rating between 1 and 5. We create 10 random $50/50$ splits of the data into a training and test set. For features we again use the $N$ most frequently occurring bigrams and for basis kernels again use $N$ rank-1 kernels, see Figure 1. The results on these dataset amplify the result from the Reuters ACQ dataset: $L_1$ regularization can negatively impact the performance for large number of kernels, while $L_2$-regularization improve the performance significantly over the baseline over the evenly-weighted sum of kernels.

## 6 Conclusion

We presented an analysis of learning kernels with ridge regression with $L_2$ regularization, including an efficient iterative algorithm. Our generalization bound suggests that with even a relatively large number of orthogonal kernels the estimation error is not significantly increased. This favorable theoretical situation is also corroborated by some of our empirical results. Our analysis was based on the stability of LKRR. We do not expect similar results to hold for $L_1$ regularization since $L_1$ typically does not ensure the same uniform stability guarantees.

## A Expression of $\Delta\mu_k$

**Lemma 1.** *For any samples $S$ and $S'$, $\Delta\mu_k$ can be expressed in terms of $\Delta v_k$ as follows:*

$$\Delta\mu_k = \Lambda\left[\frac{\Delta v_k}{\|\mathbf{v}'\|} - \frac{v_k\sum_{i=1}^p (v_i + v_i')\Delta v_i}{\|\mathbf{v}\|\|\mathbf{v}'\|(\|\mathbf{v}\| + \|\mathbf{v}'\|)}\right]. \quad (23)$$

*Proof.* By definition of $\mu_k$, we can write

$$\Delta\mu_k = \Lambda\left[\frac{v_k'}{\|\mathbf{v}'\|} - \frac{v_k}{\|\mathbf{v}\|}\right] = \Lambda\left[\frac{v_k' - v_k}{\|\mathbf{v}'\|} - \frac{v_k\|\mathbf{v}'\| - v_k\|\mathbf{v}\|}{\|\mathbf{v}\|\|\mathbf{v}'\|}\right]$$
$$= \Lambda\left[\frac{v_k' - v_k}{\|\mathbf{v}'\|} - \frac{v_k\Delta(\|\mathbf{v}\|)}{\|\mathbf{v}\|\|\mathbf{v}'\|}\right].$$

Observe that: $\Delta(\|\mathbf{v}\|) = \frac{\Delta(\|\mathbf{v}\|^2)}{\|\mathbf{v}\|+\|\mathbf{v}'\|} = \frac{\Delta(\sum_{i=1}^p v_i^2)}{\|\mathbf{v}\|+\|\mathbf{v}'\|} = \frac{\sum_{i=1}^p \Delta(v_i)(v_i+v_i')}{\|\mathbf{v}\|+\|\mathbf{v}'\|}$. Plugging in this identity in the previous one yields the statement of the lemma. ☐

## B Proof of Proposition 1

*Proof.* The terms $\Delta_K v_k$ appearing in $V_1$ have the following more explicit expression:

$$\Delta_K v_k = \Delta_K(\boldsymbol{\alpha}^\top \mathbf{K}_k(S')\boldsymbol{\alpha})$$
$$= \Delta_K(\boldsymbol{\alpha}^\top)\mathbf{K}_k(S')\boldsymbol{\alpha}' + \boldsymbol{\alpha}^\top \mathbf{K}_k(S')\Delta_K(\boldsymbol{\alpha}).$$

Thus, $V_1$ can be written as a sum $V_1 = V_{11} + V_{12}$ according to this decomposition. We shall show how $V_{12}$ is bounded, $V_{11}$ is bounded in a very similar way. In view of the expression for $V_1$ (12), and using $\mathbf{K}_k = \boldsymbol{\Phi}_k^\top\boldsymbol{\Phi}_k$, $V_{12}$ can be written as

$$V_{12} = \Lambda\sum_{k=1}^p (\Delta_K\boldsymbol{\alpha})^\top \mathbf{Z}[\boldsymbol{\Phi}_k\boldsymbol{\alpha}]^\top, \quad (24)$$

with $\mathbf{Z} = \frac{\boldsymbol{\Phi}_k^\top\boldsymbol{\Phi}_k\boldsymbol{\alpha}}{\|\mathbf{v}\|} - \frac{v_k\sum_{i=1}^p\sum_i (v_i+v_i')\boldsymbol{\Phi}_i^\top\boldsymbol{\Phi}_i\boldsymbol{\alpha}}{\|\mathbf{v}\|\|\mathbf{v}'\|(\|\mathbf{v}\|+\|\mathbf{v}'\|)}$. Using the fact that $\|\boldsymbol{\Phi}_k\boldsymbol{\alpha}\|^2 = \boldsymbol{\alpha}^\top\boldsymbol{\Phi}_k^\top\boldsymbol{\Phi}_k\boldsymbol{\alpha} = \boldsymbol{\alpha}^\top\mathbf{K}_k\boldsymbol{\alpha} = v_k$ and similarly $\|\boldsymbol{\Phi}_i\boldsymbol{\alpha}\| = v_i^{1/2}$ and assuming without loss of generality that $\|v'\| \geq \|v\|$, $V_{12}$ can be bounded by

$\Lambda\sum_{k=1}^p \|\Delta_K\boldsymbol{\alpha}\|\left(\frac{v_k}{\|\mathbf{v}\|}\|\boldsymbol{\Phi}_k\| + \frac{v_k}{\|\mathbf{v}\|}\frac{\sum_i (v_i+v_i')v_k^{1/2}v_i^{1/2}\|\boldsymbol{\Phi}_i\|}{\|\mathbf{v}'\|(\|\mathbf{v}\|+\|\mathbf{v}'\|)}\right)$. By the Cauchy-Schwarz inequality, the first sum $\sum_{k=1}^p \frac{v_k}{\|\mathbf{v}\|}\|\boldsymbol{\Phi}_k\|$ can be bounded as follows

$$\sum_{k=1}^p \frac{v_k}{\|\mathbf{v}\|}\|\boldsymbol{\Phi}_k\| \leq \frac{\|\mathbf{v}\|}{\|\mathbf{v}\|}\Big(\sum_{k=1}^p \|\boldsymbol{\Phi}_k\|^2\Big)^{1/2} \leq \sqrt{\kappa p m}, \quad (25)$$

since $\|\boldsymbol{\Phi}_k\| \leq \sqrt{\kappa m}$. The second sum is similarly simplified and bounded as follows

$$\sum_{k=1}^p \frac{v_k}{\|\mathbf{v}\|}\frac{\sum_{i=1}^p (v_i+v_i')v_k^{1/2}v_i^{1/2}\|\boldsymbol{\Phi}_i\|}{\|\mathbf{v}'\|(\|\mathbf{v}\|+\|\mathbf{v}'\|)}$$
$$\leq \left(\sum_{k=1}^p \frac{v_k^{3/2}}{\|\mathbf{v}\|}\right)\left(\sum_{i=1}^p \frac{(v_i^{3/2} + v_i'v_i^{1/2})}{\|\mathbf{v}'\|(\|\mathbf{v}\|+\|\mathbf{v}'\|)}\right)\max_i \|\boldsymbol{\Phi}_i\|.$$

In view of $\|\boldsymbol{\Phi}_i\| \leq \sqrt{\kappa m}$ for all $i$, and using multiple applications of the Cauchy-Schwarz inequality, e.g., $\sum_{k=1}^p v_k^{3/2} = \sum_{k=1}^p v_k v_k^{1/2} \leq \|\mathbf{v}\|\|\mathbf{v}\|_1^{1/2}$ and $\sum_{i=1}^p v_i'v_i^{1/2} \leq \|\mathbf{v}'\|\|\mathbf{v}\|_1^{1/2}$, the second sum is also bounded by $\sqrt{\kappa p m}$ and $\|V_{12}\| \leq 2\Lambda\sqrt{\kappa p m}\|\Delta_K\boldsymbol{\alpha}\|$. Proceeding in the same way for $V_{11}$ leads to $\|V_{11}\| \leq 2\Lambda\sqrt{\kappa p m}\|\Delta_K\boldsymbol{\alpha}\|$ and $\|V_1\| \leq 4\Lambda\sqrt{\kappa p m}\|\Delta_K\boldsymbol{\alpha}\|$. ☐

## C Proof of Proposition 2

*Proof.* The main idea of the proof is to bound $V_2$ in terms of $\Delta_S\mathbf{w}$, the difference of the weight vectors $h$ and $h'$ already bounded in the proof of Theorem 2.

By definition, $v_k = \boldsymbol{\alpha}^\top\mathbf{K}_k\boldsymbol{\alpha}$. Since $\mathbf{K}_k = \boldsymbol{\Phi}_k^\top\boldsymbol{\Phi}_k$, then $v_k = \|\mathbf{w}_k\|^2$, where $\mathbf{w}_k = \boldsymbol{\Phi}_k(S)\boldsymbol{\alpha}$. Thus, in view of (12), $V_2$ can be written as follows

$$V_2 = \Lambda\sum_{k=1}^p\left(\frac{\Delta_S\|\mathbf{w}_k\|^2}{\|\mathbf{v}'\|} - \frac{v_k\sum_i (v_i+v_i')\Delta_S\|\mathbf{w}_i\|^2}{\|\mathbf{v}\|\|\mathbf{v}'\|(\|\mathbf{v}\|+\|\mathbf{v}'\|)}\right)\mathbf{w}_k^\top.$$

We can bound $|\Delta_S\|\mathbf{w}_k\|^2|$ in terms of $\|\Delta_S\mathbf{w}_k\|$:

$$|\Delta_S\|\mathbf{w}_k\|^2| = |(\Delta_S\mathbf{w}_k)^\top\mathbf{w}_k' + \mathbf{w}_k^\top(\Delta_S\mathbf{w}_k)|$$
$$= |(\Delta_S\mathbf{w}_k)^\top(\mathbf{w}_k' + \mathbf{w}_k)| \leq \|\mathbf{w}_k' + \mathbf{w}_k\|\|\Delta_S\mathbf{w}_k\|.$$

Thus, since $\|\mathbf{w}_k\| = (\boldsymbol{\alpha}^\top\boldsymbol{\Phi}_k^\top\boldsymbol{\Phi}_k\boldsymbol{\alpha})^{1/2} \leq v_k^{1/2}$ and $\|\mathbf{w}_k'\| \leq {v'}_k^{1/2}$, $\|V_2\|$ can be bounded by

$$\|V_2\| \leq \Lambda\Bigg(\sum_{k=1}^p \frac{v_k^{1/2}(v_k^{1/2} + {v'}_k^{1/2})}{\|\mathbf{v}'\|}\|\Delta_S\mathbf{w}_k\|$$
$$+ \sum_{i=1}^p \frac{(v_i+v_i')(v_i^{1/2} + {v'}_i^{1/2})}{\|\mathbf{v}\|\,\|\mathbf{v}'\|(\|\mathbf{v}\|+\|\mathbf{v}'\|)}\|\Delta_S\mathbf{w}_i\|\,\Big\|\sum_{k=1}^p v_k\mathbf{w}_k^\top\Big\|\Bigg).$$

The first sum can be bounded as follows

$$\sum_{k=1}^{p} \frac{v_k^{1/2}(v_k^{1/2} + v'^{1/2}_k)\|\Delta_S \mathbf{w}_k\|}{\|\mathbf{v}'\|}$$
$$= \sum_{k=1}^{p} \frac{v_k + (v_k v'_k)^{1/2}}{\mu_k \|\mathbf{v}'\|} \|\Delta_S(\mu_k \mathbf{w}_k)\|$$
$$\leq \Bigg( \underbrace{\Big( \sum_{k=1}^{p} \frac{(v_k + (v_k v'_k)^{1/2})^2}{\mu_k^2 \|\mathbf{v}'\|^2} \Big)}_{F_1} \Big( \sum_{k=1}^{p} \|\Delta_S(\mu_k \mathbf{w}_k)\|^2 \Big) \Bigg)^{1/2}.$$

The first factor is bounded by a constant using multiple applications of the Cauchy-Schwarz inequality and assuming without loss of generality that $\|\mathbf{v}\| \leq \|\mathbf{v}'\|$: $F_1 = \sum_{k=1}^{p} \frac{v_k^2 + (v_k v'_k) + 2 v_k^{3/2} v'^{1/2}_k}{\mu_k^2 \|\mathbf{v}'\|^2} \leq 4$ (the calculation steps are omitted due to space). The second sum can be bounded as follows

$$\frac{\sum_i (v_i + v'_i)(v_i^{1/2} + v'^{1/2}_i)\|\Delta_S \mathbf{w}_i\|}{\|\mathbf{v}\|\|\mathbf{v}'\|(\|\mathbf{v}\| + \|\mathbf{v}'\|)} \Big\| \sum_{k=1}^{p} v_k \mathbf{w}_k^\top \Big\|$$
$$\leq \sum_{i=1}^{p} \frac{(v_i + v'_i)(v_i^{1/2} + v'^{1/2}_i)}{\mu_i \|\mathbf{v}\|\|\mathbf{v}'\|(\|\mathbf{v}\| + \|\mathbf{v}'\|)} \|\Delta_S(\mu_i \mathbf{w}_i)\| \Big\| \sum_{k=1}^{p} v_k \mathbf{w}_k \Big\|$$
$$\leq F_2 \Big[ \sum_{i=1}^{p} \|\Delta_S(\mu_i \mathbf{w}_i)\|^2 \Big]^{1/2} \Big\| \sum_{k=1}^{p} v_k \mathbf{w}_k \Big\|,$$

where $F_2 = \Big[ \sum_{i=1}^{p} \frac{(v_i + v'_i)^2 (v_i^{1/2} + v'^{1/2}_i)^2}{\|\mathbf{v}\|^2 \|\mathbf{v}'\|^2 (\|\mathbf{v}\| + \|\mathbf{v}'\|)^2} \Big]^{\frac{1}{2}}$. The numerator of $F_2$, can be bounded using $\sum_{i=1}^{p} v_i^3 \leq \|\mathbf{v}\|^3$, $\sum_{i=1}^{p} v_i^{5/2} v'^{1/2}_i \leq \|\mathbf{v}\|^{5/2}\|\mathbf{v}'\|^{1/2}$ and applications of the Cauchy-Schwarz inequality such as $\sum_{i=1}^{p}(v_i + v'_i)^2 (v_i^{1/2} + v'^{1/2}_i)^2 \leq (\|\mathbf{v}\| + \|\mathbf{v}'\|)^2 (\|\mathbf{v}\|^{1/2} + \|\mathbf{v}'\|^{1/2})^2$. The intermediate steps are omitted due to space. This leads to $F_2 \leq \frac{\|\mathbf{v}\|^{1/2} + \|\mathbf{v}'\|^{1/2}}{\|\mathbf{v}\|\|\mathbf{v}'\|}$ and

$$\|V_2\| \leq 2\Lambda \Bigg( 1 + \frac{\|\mathbf{v}\|^{1/2} + \|\mathbf{v}'\|^{1/2}}{2\|\mathbf{v}\|\,\|\mathbf{v}'\|} \Big\| \sum_{k=1}^{p} v_k \mathbf{w}_k \Big\| \Bigg) F_3,$$

with $F_3 = \big( \sum_{k=1}^{p} \|\Delta_S \mu_k \mathbf{w}_k\|^2 \big)^{1/2}$. If the feature vectors $\mathbf{w}_k$ are orthogonal, that is $\mathbf{w}_k^\top \mathbf{w}_{k'} = 0$ for $k \neq k'$ (which holds in particular if $\Phi_k(x_i)^\top \Phi_{k'}(x_i) = 0$ for $k \neq k'$ and $i = 1, \ldots, m$), then $F_3 = \|\Delta_S \mathbf{w}\|$ and $\big\| \sum_{k=1}^{p} v_k \mathbf{w}_k \big\|^2 = \sum_{k=1}^{p} v_k^2 \mathbf{w}_k^\top \mathbf{w}_k = \sum_{k=1}^{p} v_k^3 \leq \|\mathbf{v}\|^3$. Thus, using the bound on $\|\Delta_S \mathbf{w}\|$ from the proof of Theorem 2 yields

$$\|V_2\| \leq 2\Lambda \Bigg( 1 + \frac{\|\mathbf{v}\|^{1/2} + \|\mathbf{v}'\|^{1/2}}{2\|\mathbf{v}\|\|\mathbf{v}'\|} \|\mathbf{v}\|^{3/2} \Bigg) \|\Delta_S \mathbf{w}\|$$
$$\leq 4\Lambda \|\Delta_S \mathbf{w}\| \leq \frac{4\Lambda M}{\lambda_{\min} + \lambda_0 m} \leq \frac{4\Lambda M}{\lambda_0 m}. \qquad \square$$

## References

Argyriou, A., Hauser, R., Micchelli, C., & Pontil, M. (2006). A DC-programming algorithm for kernel selection. *ICML*.

Argyriou, A., Micchelli, C., & Pontil, M. (2005). Learning convex combinations of continuously parameterized basic kernels. *COLT*.

Bach, F. (2008). Exploring large feature spaces with hierarchical multiple kernel learning. *NIPS*.

Blitzer, J., Dredze, M., & Pereira, F. (2007). Biographies, Bollywood, Boom-boxes and Blenders: Domain Adaptation for Sentiment Classification. *Association for Computational Linguistics*.

Boser, B., Guyon, I., & Vapnik, V. (1992). A training algorithm for optimal margin classifiers. *COLT*.

Bousquet, O., & Elisseeff, A. (2002). Stability and generalization. *JMLR*, *2*.

Bousquet, O., & Herrmann, D. J. L. (2002). On the complexity of learning the kernel matrix. *NIPS*.

Cortes, C., Mohri, M., & Rostamizadeh, A. (2008). Learning sequence kernels. *MLSP*.

Cortes, C., & Vapnik, V. (1995). Support-Vector Networks. *Machine Learning*, *20*.

Jebara, T. (2004). Multi-task feature and kernel selection for SVMs. *ICML*.

Lanckriet, G., Cristianini, N., Bartlett, P., Ghaoui, L. E., & Jordan, M. (2004). Learning the kernel matrix with semidefinite programming. *JMLR*, *5*.

Lewis, D. P., Jebara, T., & Noble, W. S. (2006). Nonstationary kernel combination. *ICML*.

Micchelli, C., & Pontil, M. (2005). Learning the kernel function via regularization. *JMLR*, *6*.

Ong, C. S., Smola, A., & Williamson, R. (2005). Learning the kernel with hyperkernels. *JMLR*, *6*.

Saunders, C., Gammerman, A., & Vovk, V. (1998). Ridge Regression Learning Algorithm in Dual Variables. *ICML*.

Schölkopf, B., & Smola, A. (2002). *Learning with kernels*. MIT Press: Cambridge, MA.

Shawe-Taylor, J., & Cristianini, N. (2004). *Kernel methods for pattern analysis*. Cambridge Univ. Press.

Srebro, N., & Ben-David, S. (2006). Learning bounds for support vector machines with learned kernels. *COLT*.

Vapnik, V. N. (1998). *Statistical learning theory*. John Wiley & Sons.

von Neumann, J. (1937). Uber ein ökonomisches Gleichungssystem . *Ergebn. Math. Kolloq. Wein 8*.

Zien, A., & Ong, C. S. (2007). Multiclass multiple kernel learning. *ICML*.